\documentclass[]{article}
\usepackage[T1]{fontenc}
\usepackage[utf8]{inputenc}
\usepackage{textcomp}
\usepackage{lmodern}
\usepackage{amsmath,amssymb}
\usepackage[margin=1in]{geometry}
\usepackage{microtype}
\UseMicrotypeSet[protrusion]{basicmath}
\usepackage{parskip}
\usepackage{xcolor}
\usepackage{longtable,booktabs,array}
\usepackage{calc}
\usepackage{etoolbox}
\makeatletter
\patchcmd\longtable{\par}{\if@noskipsec\mbox{}\fi\par}{}{}
\makeatother
\usepackage{footnote}
\makesavenoteenv{longtable}
\usepackage{graphicx}
\makeatletter
\def\maxwidth{\ifdim\Gin@nat@width>\linewidth\linewidth\else\Gin@nat@width\fi}
\def\maxheight{\ifdim\Gin@nat@height>\textheight\textheight\else\Gin@nat@height\fi}
\makeatother
\setkeys{Gin}{width=\maxwidth,height=\maxheight,keepaspectratio}
\makeatletter
\def\fps@figure{htbp}
\makeatother
\usepackage{soul}
\usepackage{bookmark}
\usepackage{xurl}
\hypersetup{
  pdftitle={The Feasibility of Training Sovereign Language Models in the Global South: A Study of Brazil and Mexico},
  hidelinks,
  pdfcreator={Monica Ulloa},
  breaklinks=true
}

\usepackage{breakurl}
\usepackage{authblk}

\title{\phantomsection\label{_4mh9yiwnsi6q}{}The Feasibility of Training
Sovereign Language Models in the Global South: A Study of Brazil and
Mexico}
\author[1,2]{Sandra Malagon}
\author[1, 2]{Monica A. Ulloa Ruiz}
\author[1]{Tatiana Elizabeth Sandoval Plaza}
\author[1]{Gabriel Rafael Rosario Bolívar}
\author[1]{Valentina García Mesa}
\author[1]{Ivanna Alvarado Morales}
\affil[1]{Carreras con Impacto}
\affil[2]{Aixo}
\date{}

\begin{document}
\maketitle

\begin{abstract}
The rapid escalation of computational requirements for training large-scale language models has reinforced structural asymmetries between high-capacity jurisdictions and countries in the Global South. This paper examines the technical and fiscal feasibility of sovereign-scale language model training in Brazil and Mexico under conditions of constrained hardware access, energy availability, and fiscal ceilings. Using a dual-axis design that varies accelerator generation (NVIDIA H100 vs. A100) and training duration (90 vs. 150 days), we estimate compute demand, energy consumption, capital expenditures, and regulatory compatibility for the training of a 10-trillion-token model. Our findings show that while all configurations remain below export-control and electrical infrastructure thresholds, fiscal viability is determined by hardware efficiency. H100-based scenarios achieve training feasibility at a total cost of 8–14 million USD, while A100 deployments require 19–32 million USD due to higher energy and hardware demand. We argue that extending training timelines should be treated as a policy lever to mitigate hardware constraints, enabling the production of usable, auditable, and locally aligned models without competing at the global frontier. This study contributes to the discourse on AI compute governance and technological sovereignty by highlighting context-sensitive strategies that allow middle-income countries to establish sustainable and strategically sufficient AI capabilities.
\end{abstract}

\section{1. Introduction}\label{introduction}

The rapid development of frontier-scale language models has accelerated
global demand for compute---but also reinforced longstanding
inequalities in technological access. Training these systems requires
thousands of high-end GPUs running uninterrupted for weeks, supported by
robust electrical infrastructure, cooling systems, and data centre
engineering. As a result, computational capacity is increasingly
concentrated in a handful of jurisdictions, leading to what researchers
have termed a new ``GPU North--South divide'' (Sastry et al., 2024).
Countries outside this core are not only less able to train frontier
models, but may also struggle to adapt or audit pre-trained systems in
ways that reflect local linguistic, legal, and institutional priorities.

At the same time, energy availability has emerged as a strategic
constraint in the deployment of AI infrastructure---particularly in
countries of the Global South, where electrical grids may be less robust
or subject to capacity and reliability limits. The International Energy
Agency projects that electricity demand from AI and data centres could
exceed 1,000 TWh annually by 2026, a figure comparable to the total
power consumption of mid-sized economies (IEA, 2024). A single GPT-3
training run consumed more than 1.3 GWh (Patterson et al., 2021), and
inference demands have since grown even larger and more geographically
distributed. In many regions, access to high-end accelerators such as
the NVIDIA H100 is limited by export controls (BIS, 2025) or market
constraints---leaving prior-generation chips like the A100, still
available through secondary markets and data centre liquidations, as the
most viable alternative. In such settings, both fiscal cost and
electricity availability become binding considerations, since older GPUs
incur higher energy and hardware costs per unit of compute.

This paper investigates the feasibility of training sovereign
large-scale language models under these dual constraints. We ask whether
countries like Brazil and Mexico could produce usable, non-frontier
models by trading hardware efficiency for longer training cycles---using
legacy GPUs over extended periods to offset compute limitations while
remaining within infrastructure and budgetary bounds.

Specifically, we compare four infrastructure scenarios for training a
10-trillion-token model (comparable in scale to DeepSeek
671B\footnote{We adopt a compute profile aligned with DeepSeek-V3---a
  recent large-scale model trained using a Mixture-of-Experts (MoE)
  architecture---due to its technical transparency, use of publicly
  documented infrastructure, and relevance as a strategic benchmark for
  middle-power jurisdictions. This scale was selected to reflect a
  credible frontier-level model capable of supporting general-purpose
  applications in science, government, education, and national
  infrastructure.}), using both H100 and A100 clusters across 90-day and
150-day training windows. By varying hardware type and training duration
independently, this dual-axis design enables us to disentangle two
strategic constraints: (1) limited access to high-end GPUs under export
restrictions, and (2) fiscal and energy costs associated with sustained
electrical load. Unlike prior studies that focus exclusively on
optimized frontier-scale H100 clusters (Sastry et al., 2024), our
analysis explicitly models energy consumption, capital expenditure, and
regulatory feasibility across a range of configurations---drawing on
component-level thermal design power (TDP), PUE-adjusted cooling
requirements, and jurisdiction-specific hardware pricing.

Our results offer three key contributions:

\begin{itemize}
\item
  We show that---even with constrained infrastructure---training a
  large, usable model remains technically and economically viable for
  Tier 2 countries.
\item
  We demonstrate that A100-based deployments are less fiscally viable,
  as their lower efficiency drives up both electricity consumption and
  capital expenditures.
\item
  We argue that training time should be treated as a policy lever, not
  merely a technical parameter: by extending training windows, countries
  in the Global South can adopt alternative compute strategies without
  requiring immediate access to frontier accelerators.
\end{itemize}

In doing so, we contribute to the broader discourse on compute
governance, energy-aware AI planning, and technological sovereignty in
the Global South. Rather than focusing on frontier competitiveness, our
analysis highlights viable, context-sensitive strategies for training
models that are ``good enough''---usable, auditable, and locally
aligned---even in the absence of cutting-edge hardware.

\section{2. Methodology}\label{methodology}

\subsection{2.1 Training Model
Assumptions}\label{training-model-assumptions}

We assume a total training compute budget of 3.0×10\^{}24 FLOPs, based
on a benchmark consistent with DeepSeek‑V3---a 671B-parameter
Mixture-of-Experts model trained over 14.8 trillion tokens with
approximately 37 billion active parameters per forward-backward pass.
This estimate aligns with standard Transformer training approximations
and the reporting methodology used by Epoch AI (2024), which tracks
models trained at scales of 10\^{}23 FLOPs.

This compute budget serves as a fixed input across all infrastructure
scenarios evaluated in this study and reflects the minimum scale
necessary to train a strategically sufficient model: one capable of
supporting sovereign, institutional, or regional deployments, but
falling below current definitions of \emph{frontier AI} in terms of
general-purpose capability or systemic risk.

We define a strategically sufficient model as a large-scale language
model that that does not meet the definition of frontier AI---described
by the UK Government (DSIT, 2025) as highly capable general-purpose
models that can perform a wide variety of tasks and match or exceed the
capabilities of today's most advanced systems---but achieves the level
of performance required to support public functions like education,
legal reasoning, administrative automation, and local language
alignment. These models are deployable on constrained infrastructure,
auditable under public governance frameworks, and adaptable to national
or institutional priorities.

For example, DeepSeek‑V3 was trained using approximately 10× less
compute than GPT‑4, which Epoch AI (2024) estimates at 1.8×10\^{}25 to
3.0×10\^{}25 FLOPs. Despite this, it achieves an aggregate quality score
of 80.0 on ArtificialAnalysis.ai (2025), outperforming all open-weight
models and scoring within \textasciitilde10 percentage points of GPT‑4o.

\subsection{2.2 Hardware and Training Duration
Scenarios}\label{hardware-and-training-duration-scenarios}

We evaluate four infrastructure configurations combining two accelerator
classes---NVIDIA H100 and NVIDIA A100---with two training schedules: 90
days and 150 days. These scenarios are designed to capture key
trade-offs between hardware generation, energy efficiency, and
deployment feasibility.

The modeling assumptions are as follows:

\begin{itemize}
\item
  H100 GPUs are assigned a peak throughput of 2,000 TFLOPs using FP8
  precision via NVIDIA's Transformer Engine (NVIDIA).
\item
  A100 GPUs are assigned a peak throughput of 312 TFLOPs under FP16
  precision, as specified in NVIDIA's official architectural
  documentation (NVIDIA).
\item
  A Model FLOP Utilization (MFU) factor of 0.552 is applied to all
  scenarios to account for real-world training inefficiencies,
  consistent with the empirical estimates compiled by Sevilla et al.
  (2022).\\
  The total number of GPUs required in each scenario is computed using:
\end{itemize}

\begin{align*}
  \text{GPUs} &= \frac{\text{Total FLOPs}}{\text{Training time (s)} \times
\text{Peak TFLOPs per GPU} \times 0.552}
\end{align*}

Where training time is expressed in seconds over 90 or 150 days. This
results in GPU requirements ranging from \textasciitilde350 H100s (in
the 150-day scenario) to over 2,200 A100s (in the 90-day scenario),
illustrating the impact of both hardware efficiency and extended
training schedules on infrastructure needs. Detailed breakdowns of GPU
requirements for each scenario are provided in Annex 1.

\subsection{2.3 Estimation of Resources}\label{estimation-of-resources}

\subsubsection{Energy consumption (MWh)}\label{energy-consumption-mwh}

Total energy consumption is calculated as the product of training
duration, GPU thermal design power (TDP), and total GPU count, adjusted
by a Power Usage Effectiveness (PUE) of 1.3 to reflect datacenter
overhead. We assume:

\begin{itemize}
\item
  700\,W TDP per H100 GPU
\item
  400\,W TDP per A100 GPU
\end{itemize}

\begin{align*}
  \text{Energy (MWh)}&=\frac{\text{NGPU}\times\text{TDPadj}\times\text{PUE}\times\text{D}\times 24}{10^6}
\end{align*}

These estimates yield energy requirements between 0.3 and 3.3 GWh
depending on hardware type and training schedule. Longer schedules and
newer hardware significantly reduce cumulative consumption.

\subsubsection{Peak electrical load (MW)}\label{peak-electrical-load-mw}

We estimate peak demand by assuming full simultaneous usage of all GPUs
in a given configuration:

\begin{align*}
  \text{Peak Load (MW)} &= \frac{N_{\text{GPU}} \times \text{TDP}_{\text{adj}} \times \text{PUE}}{10^{6}}
\end{align*}

Peak load ranges from 0.41 MW (H100, 150 days) to 1.49 MW (A100, 90
days), remaining within the envelope of medium-voltage distribution
infrastructure in most Tier 2 urban industrial parks.

\subsubsection{\texorpdfstring{Capital Expenditures (CAPEX)\\
Hardware prices are set
at:}{Capital Expenditures (CAPEX) Hardware prices are set at:}}\label{capital-expenditures-capex-hardware-prices-are-set-at}

\begin{itemize}
\item
  33,000\,USD per H100 GPU
\item
  12,000\,USD per A100 GPU
\end{itemize}

A 30\% integration overhead is applied to account for additional
components (CPUs, memory, SSDs, NICs, chassis, etc.). Tariff assumptions
are country-specific:

\begin{itemize}
\item
  Brazil: 16\% import duty on GPU hardware (International Trade
  Administration, 2021)
\item
  Mexico: 0\% import duty, consistent with digital infrastructure
  exemptions (Gobierno de México, 2023)
\end{itemize}

\begin{align*}
  \text{CAPEX}_{\text{USD}} = \text{Number of GPUs} \times
\text{Price per GPU} \times 1.30 \times (1 + \text{Import Tariff})
\end{align*}

CAPEX dominates total system cost in all scenarios, ranging from 1.06
million USD to 13.49 million USD depending on configuration.

\subsubsection{Operating Expenditures
(OPEX)}\label{operating-expenditures-opex}

Electricity costs are based on the total MWh consumed per scenario and
the applicable industrial rates:

\begin{itemize}
\item
  Brazil: 110\,USD/MWh (ANEEL, 2024)
\item
  Mexico: 88\,USD/MWh (Comisión Federal de electricidad, 2025)
\end{itemize}

\begin{align*}
  \text{OPEX}_{\text{USD}} &= \text{Energy (MWh)}\times \text{Tariff (USD/MWh)}
\end{align*}

For instance, in the \emph{H100 · 90d} scenario, Brazil consumed 893 MWh
at 110 USD/MWh, yielding

\begin{align*}
  \text{OPEX}_{\text{USD}} &= 893 \times 110 = 98.230 \text{ USD } (0.098 M)
\end{align*}

while in Mexico the same energy use at 88 USD/MWh resulted in

\begin{align*}
  \text{OPEX}_{\text{USD}} &= 893 \times 88 = 78.584 \text{ USD } (0.079 M)
\end{align*}

In the \emph{A100 · 90d} scenario, energy consumption was 3,271 MWh;
applying the tariffs gives 359,799 USD (0.360 M) for Brazil and 287,839
USD (0.288 M) for Mexico.

Across all scenarios, OPEX remains below 0.4 million USD---well under
5\% of total cost---highlighting that training expenditure is
overwhelmingly driven by hardware acquisition rather than energy use.
Full cost breakdowns by country and scenario are provided in Annex 1.

\subsection{2.4 Feasibility Constraints and Evaluation
Framework}\label{feasibility-constraints-and-evaluation-framework}

\subsubsection{Export-control ceiling ($\le$ 50,000
GPUs)}\label{export-control-ceiling-50000-gpus}

We adopt a GPU cap of 50,000 units per country, in line with recent
export controls issued by the U.S. Bureau of Industry and Security (BIS,
2025). These restrictions apply to high-end accelerators such as the
H100 and A100, and are already enforced in key regions like China and
the Middle East. This threshold reflects growing global scrutiny over
compute flows and offers a conservative yet actionable boundary for
national-scale deployment.

\subsubsection{Electrical infrastructure
limit}\label{electrical-infrastructure-limit}

We establish a dual-tiered infrastructure constraint:

\begin{itemize}
\item
  A maximum peak load of 10 megawatts (MW) is adopted as a hard upper
  limit for urban deployments without high-voltage interconnection.
  Loads beyond this level typically require dedicated substations and
  specialized grid upgrades (SENER, 2023; Enel-SP, 2023).
\item
  More critically, we identify 1\,MW as the practical threshold for
  deployment without additional permitting or physical infrastructure
  upgrades in urban industrial parks and university zones. This is based
  on capacity studies in Querétaro Comisión Federal de Electricidad
  (2024) and São Paulo, where 1--2\,MW loads are generally supportable
  within existing 13--34\,kV distribution networks.
\end{itemize}

1\,MW often triggers additional barriers: transformer reinforcement,
redundant feeder lines, and, in some jurisdictions, environmental impact
assessments or public utility filings. While not insurmountable, these
burdens may delay or limit deployments by public institutions with
restricted engineering or legal capacity.

\subsubsection{Fiscal feasibility cap}\label{fiscal-feasibility-cap}

\begin{quote}
We adopt 52 million USD as the fiscal ceiling for sovereign AI
infrastructure projects, inclusive of CAPEX and first-year OPEX. This
threshold was chosen taking as reference the first-year investment
spending of a recent digital infrastructure project, Internet para Todos
México (CFE, 2024).
\end{quote}

\subsection{2.5 Case Selection: Querétaro and São
Paulo}\label{case-selection-queruxe9taro-and-suxe3o-paulo}

To evaluate the real-world feasibility of sovereign AI infrastructure
deployment, we anchor our modeling in two reference locations:
Querétaro, in central Mexico, and São Paulo, in southeastern Brazil.
Both regions offer a realistic testbed for subnational AI clusters in
upper-middle-income countries, combining medium-voltage grid access,
industrial concentration, and public infrastructure alignment.

In Querétaro, national planning documents identify the municipalities of
El Marqués and Corregidora as focal points for industrial growth and
medium-voltage energy reinforcement. The Programa de Desarrollo del
Sistema Eléctrico Nacional (PRODESEN) outlines multiple investment
projects to expand substation capacity and ensure energy redundancy for
new industrial zones, many of which operate on 13--34\,kV distribution
systems (SENER, 2023). These networks can typically support sustained
loads in the 3--5\,MW range without requiring high-voltage
interconnection. Within this context, AI training scenarios consuming up
to 100 MWh/day and drawing peak loads below 5\,MW remain within feasible
deployment bounds for sovereign compute clusters.

Similarly, in São Paulo, public infrastructure data from the Operador
Nacional do Sistema Elétrico (ONS) shows that industrial districts in
Barueri, Campinas, and Guarulhos are routinely supplied by
medium-voltage feeders capable of managing 5--15\,MW loads without
dedicated substations (ONS, 2009). These areas have historically hosted
datacenters and mission-critical services and are identified as priority
regions under Brazil's Plano Nacional de Conectividade (MCTI, 2022).
Grid conditions in these zones are compatible with the types of training
configurations modeled in this study, especially those remaining under
5\,MW of peak demand.

While both sites present viable conditions for AI deployment at moderate
scale, configurations exceeding 10\,MW---used in some comparative
scenarios---would likely trigger additional permitting, infrastructure
upgrades, or grid reinforcement, making them significantly more complex
and less immediately viable.

\section{3. Results}\label{results}

\subsection{3.1 Hardware Requirements}\label{hardware-requirements}

The scale of compute required varies substantially across scenarios,
driven by both hardware generation and training duration. In the most
demanding case---A100s over 90 days---the cluster exceeds 2,200 units,
while the most efficient configuration---H100s over 150 days---requires
fewer than 400 units. This six-fold spread underscores the central role
of accelerator choice and schedule design in determining feasibility.

The efficiency differential is especially pronounced across generations.
With a peak throughput of 2,000 TFLOPs, the H100 reduces total GPU
demand by an order of magnitude relative to the A100 at 312 TFLOPs.
Extending training from 90 to 150 days compounds this effect, lowering
requirements by roughly 40 percent across both hardware classes. These
dynamics set the practical envelope for sovereign-scale deployments, as
summarized in Figure 1.

\begin{figure*}
  \centering
  \includegraphics[width=0.8\textwidth]{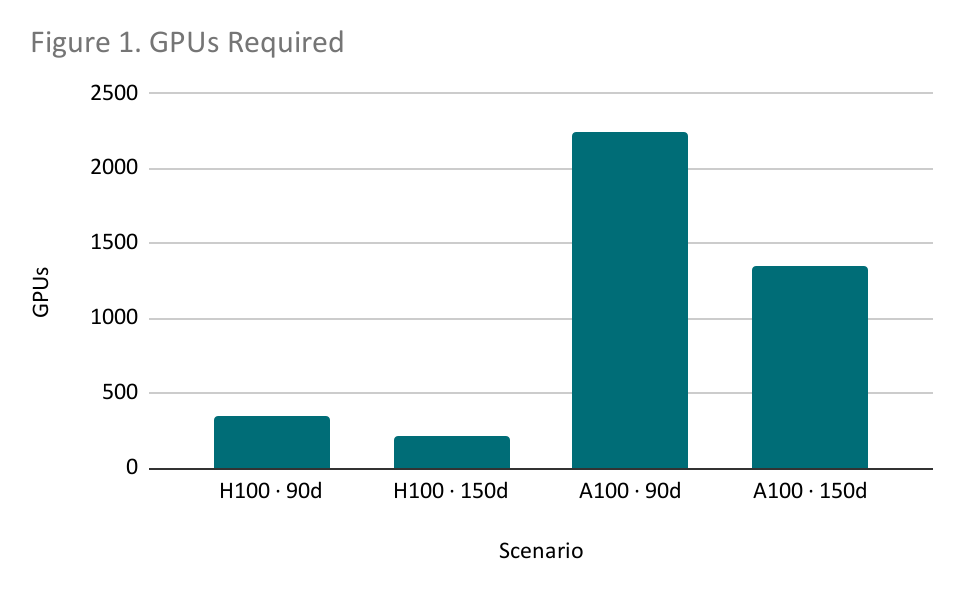}
  \caption{GPU requirements by hardware class and training duration.}
  \label{fig:figure_1}
\end{figure*}

\subsection{3.2 Energy Consumption}\label{energy-consumption}

Energy requirements remain within a moderate range across all scenarios.
The most efficient configuration (H100--150d) consumes around 0.3 GWh
over the course of training, while the most demanding (A100--90d)
reaches 3.3 GWh. These values, detailed in Figure 2, correspond to daily
averages that fall comfortably below the thresholds typically associated
with high-voltage interconnection. In practice, they remain compatible
with the supply conditions of medium-voltage industrial parks.

\begin{figure*}
  \centering
  \includegraphics[width=0.8\textwidth]{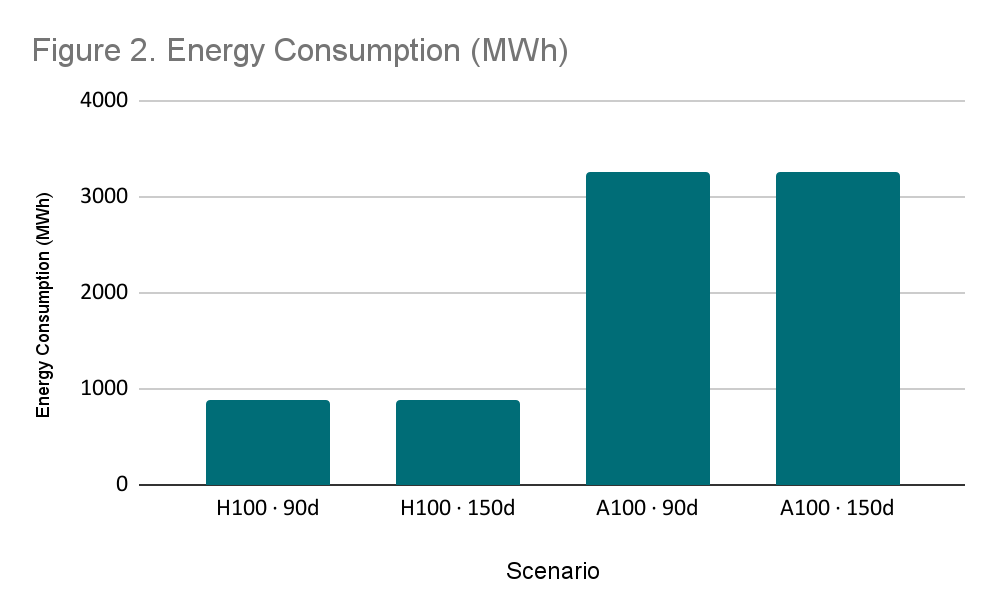}
  \caption{Total energy consumption (GWh).}
  \label{fig:figure_2}
\end{figure*}

Patterns in peak electrical load reinforce this conclusion. As reported
in Figure 3, the largest configuration (A100--90d) requires 1.49 MW,
compared to only 0.41 MW in the H100--150d case. Both figures remain
under the 10 MW ceiling used in our feasibility framework. Yet
configurations that approach or cross the 1 MW practical threshold would
likely face additional permitting and infrastructure requirements in
urban deployments, introducing a critical distinction between ``formally
viable'' and ``easily deployable.''

\begin{figure*}
  \centering
  \includegraphics[width=0.8\textwidth]{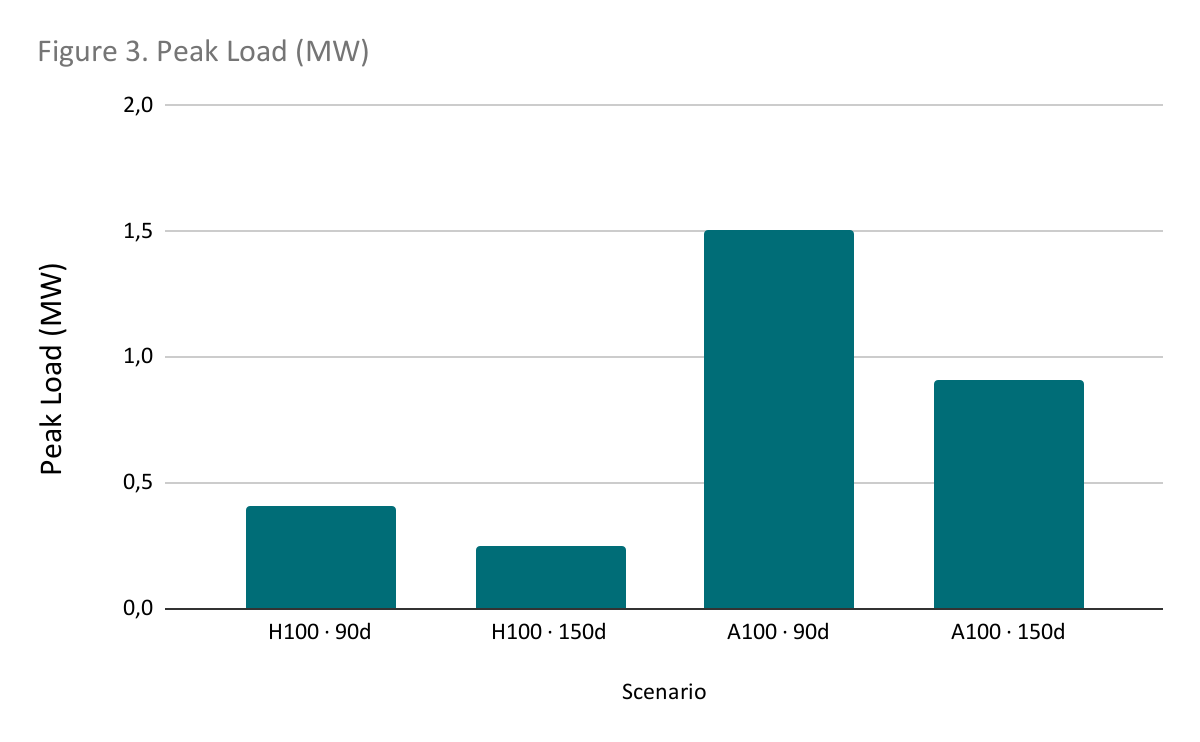}
  \caption{Peak electrical load (MW).}
  \label{fig:figure_2}
\end{figure*}

\subsection{3.3 Costs}\label{costs}

Capital expenditures represent the dominant component of training costs
across all scenarios. Hardware requirements vary from 8.3 to 13.8
million USD in the H100 configurations and from 19.3 to 32.3 million USD
in the A100 configurations. As reported in Table 1, these values already
incorporate a 30 percent integration overhead and country-specific
tariff regimes. The contrast between Mexico's 0 (Gobierno de México,
2023; BIS,2025) percent import duty and Brazil's 16 percent duty is
material, raising Brazilian deployments by several million dollars at
scale.

By comparison, operating expenditures remain marginal. Electricity costs
fall between 0.08 and 0.36 million USD, depending on scenario and
location (CFE, 2025; ANEEL, 2024). Table 1 also shows that even under
the most energy-intensive configuration, OPEX accounts for less than
five percent of total training costs. This confirms that fiscal
feasibility hinges overwhelmingly on accelerator generation and tariff
regimes rather than energy pricing.

\begin{longtable}[]{@{}
  >{\raggedright\arraybackslash}p{(\columnwidth - 8\tabcolsep) * \real{0.3020}}
  >{\raggedright\arraybackslash}p{(\columnwidth - 8\tabcolsep) * \real{0.1980}}
  >{\raggedright\arraybackslash}p{(\columnwidth - 8\tabcolsep) * \real{0.1800}}
  >{\raggedright\arraybackslash}p{(\columnwidth - 8\tabcolsep) * \real{0.2020}}
  >{\raggedright\arraybackslash}p{(\columnwidth - 8\tabcolsep) * \real{0.1180}}@{}}
\toprule\noalign{}
\begin{minipage}[b]{\linewidth}\raggedright
Scenario
\end{minipage} & \begin{minipage}[b]{\linewidth}\raggedright
Country
\end{minipage} & \begin{minipage}[b]{\linewidth}\raggedright
CAPEX (M USD)
\end{minipage} & \begin{minipage}[b]{\linewidth}\raggedright
OPEX (M USD)
\end{minipage} & \begin{minipage}[b]{\linewidth}\raggedright
TOTAL
\end{minipage} \\
\begin{minipage}[b]{\linewidth}\raggedright
H100 · 90d
\end{minipage} & \begin{minipage}[b]{\linewidth}\raggedright
Brazil
\end{minipage} & \begin{minipage}[b]{\linewidth}\raggedright
13,78
\end{minipage} & \begin{minipage}[b]{\linewidth}\raggedright
0,10
\end{minipage} & \begin{minipage}[b]{\linewidth}\raggedright
13,88
\end{minipage} \\
\begin{minipage}[b]{\linewidth}\raggedright
H100 · 90d
\end{minipage} & \begin{minipage}[b]{\linewidth}\raggedright
Mexico
\end{minipage} & \begin{minipage}[b]{\linewidth}\raggedright
13,82
\end{minipage} & \begin{minipage}[b]{\linewidth}\raggedright
0,08
\end{minipage} & \begin{minipage}[b]{\linewidth}\raggedright
13,90
\end{minipage} \\
\begin{minipage}[b]{\linewidth}\raggedright
H100 · 150d
\end{minipage} & \begin{minipage}[b]{\linewidth}\raggedright
Brazil
\end{minipage} & \begin{minipage}[b]{\linewidth}\raggedright
8,28
\end{minipage} & \begin{minipage}[b]{\linewidth}\raggedright
0,10
\end{minipage} & \begin{minipage}[b]{\linewidth}\raggedright
8,37
\end{minipage} \\
\begin{minipage}[b]{\linewidth}\raggedright
H100 · 150d
\end{minipage} & \begin{minipage}[b]{\linewidth}\raggedright
Mexico
\end{minipage} & \begin{minipage}[b]{\linewidth}\raggedright
8,32
\end{minipage} & \begin{minipage}[b]{\linewidth}\raggedright
0,08
\end{minipage} & \begin{minipage}[b]{\linewidth}\raggedright
8,39
\end{minipage} \\
\begin{minipage}[b]{\linewidth}\raggedright
A100 · 90d
\end{minipage} & \begin{minipage}[b]{\linewidth}\raggedright
Brazil
\end{minipage} & \begin{minipage}[b]{\linewidth}\raggedright
32,24
\end{minipage} & \begin{minipage}[b]{\linewidth}\raggedright
0,36
\end{minipage} & \begin{minipage}[b]{\linewidth}\raggedright
32,60
\end{minipage} \\
\begin{minipage}[b]{\linewidth}\raggedright
A100 · 90d
\end{minipage} & \begin{minipage}[b]{\linewidth}\raggedright
Mexico
\end{minipage} & \begin{minipage}[b]{\linewidth}\raggedright
32,26
\end{minipage} & \begin{minipage}[b]{\linewidth}\raggedright
0,29
\end{minipage} & \begin{minipage}[b]{\linewidth}\raggedright
32,54
\end{minipage} \\
\begin{minipage}[b]{\linewidth}\raggedright
A100 · 150d
\end{minipage} & \begin{minipage}[b]{\linewidth}\raggedright
Brazil
\end{minipage} & \begin{minipage}[b]{\linewidth}\raggedright
19,34
\end{minipage} & \begin{minipage}[b]{\linewidth}\raggedright
0,36
\end{minipage} & \begin{minipage}[b]{\linewidth}\raggedright
19,70
\end{minipage} \\
\begin{minipage}[b]{\linewidth}\raggedright
A100 · 150d
\end{minipage} & \begin{minipage}[b]{\linewidth}\raggedright
Mexico
\end{minipage} & \begin{minipage}[b]{\linewidth}\raggedright
19,35
\end{minipage} & \begin{minipage}[b]{\linewidth}\raggedright
0,29
\end{minipage} & \begin{minipage}[b]{\linewidth}\raggedright
19,64
\end{minipage} \\
\midrule\noalign{}
\endhead
\bottomrule\noalign{}
\endlastfoot
\end{longtable}

\emph{Table 1: Capital and operating expenditures (M USD) by hardware
class, training duration, and country.}

\subsection{3.4 Feasibility Assessment}\label{feasibility-assessment}

Each configuration was evaluated against the three feasibility
thresholds established by export controls, electrical infrastructure,
and fiscal ceilings. The results, show that export and power constraints
are formally non-binding: all scenarios remain well below the 50,000-GPU
export cap and the 10 MW power ceiling.

The fiscal constraint is ultimately driven by energy inefficiency: A100s
require many more units, which inflates both electricity demand and
capital outlays. Because A100s deliver substantially lower throughput
per unit than H100s, they require far more accelerators to reach the
same compute target. This larger fleet increases both cumulative
electricity demand and capital outlays, pushing total costs into the
19--32 million.

By contrast, the H100 configurations achieve the same training target
with fewer units, lower energy consumption, and total costs between 8
and 14 million USD. This makes them the most cost-effective
configurations that satisfy the full set of feasibility criteria.

\section{4. Discussion}\label{discussion}

\subsection{4.1 Sovereign Compute Is Technically
Feasible}\label{sovereign-compute-is-technically-feasible}

The results of this study demonstrate that sovereign-scale training of
\emph{usable} large language models is technically within reach for
middle-income countries. Using conservative assumptions, H100-based
configurations require between 8 and 14 million USD, while A100-based
configurations range from 19 to 32 million USD. These costs are orders
of magnitude below those associated with \emph{frontier} AI systems,
which typically demand hundreds of millions to billions of dollars in
compute and energy resources (Sevilla et al., 2022; Sastry et al.,
2024).

This distinction between \emph{usable} and \emph{frontier} models is
central for the Global South. Usable models may not match the absolute
performance of cutting-edge systems, but they are sufficient to support
key applications, enhance local capacity, and enable transparency and
auditing. Earlier analyses also emphasize that the relevant policy
question is not whether Global South actors can match frontier
capabilities, but whether they can build models that are strategically
sufficient for governance and institutional needs. Our findings confirm
that, under efficient hardware and moderate training timelines, such
deployments fall within realistic fiscal envelopes for sovereign
investment programs.

\subsection{4.2 Efficiency Is the Decisive
Variable}\label{efficiency-is-the-decisive-variable}

The decisive factor separating scenarios is hardware efficiency. A100
accelerators deliver roughly six times fewer FLOPs per unit than H100s,
which means that training the same model requires an order of magnitude
more devices. While individual A100s are cheaper, the need for
additional units increases both capital expenditures and electricity
consumption, raising total costs to 19--32 million USD. By contrast,
H100 scenarios achieve the same training target with far fewer devices,
keeping total costs within 8--14 million USD.

This pattern is consistent with prior analyses. Sevilla et al. (2022)
show that generational improvements in accelerator efficiency---both
FLOPs per dollar and FLOPs per watt---are the primary drivers of the
declining cost of training. Sastry et al. (2024) similarly note that
efficiency gains, not absolute budgets, determine whether national
actors can feasibly deploy sovereign compute clusters.

Our results reinforce this conclusion: fiscal viability in the Global
South depends less on headline budgets than on the choice of hardware
generation. While both H100- and A100-based deployments fall within a
feasible range, the former are more affordable and cost-effective,
placing sovereign training projects inside a realistic fiscal envelope.
Efficiency thus plays a dual role: it lowers overall outlays and keeps
deployments compatible with existing medium-voltage infrastructure,
delaying---but not eliminating---the need for substantial grid upgrades.

\subsection{4.3 Export Controls Are Not the
Bottleneck}\label{export-controls-are-not-the-bottleneck}

Export controls do not present a binding constraint for the scenarios
evaluated in this study. All configurations fall well below the
50,000-GPU ceiling that has been proposed as a reference point in recent
policy debates (Sastry et al., 2024). From a regulatory perspective,
middle-income countries would therefore be able to import the required
hardware without triggering export restrictions.

The more significant bottlenecks lie elsewhere. First, hardware
efficiency determines whether costs remain within a realistic fiscal
envelope. Second, while current deployments fit within medium-voltage
industrial infrastructure, any attempt to scale toward larger clusters
would require substantial upgrades to national electricity grids. As
noted in national planning documents such as PRODESEN (SENER,2023) and
EPE (Empresa de Pesquisa Energética, 2023) long-term capacity
projections (Brazil), sustained sovereign compute will ultimately depend
on aligning AI ambitions with infrastructure investment.

\subsection{4.4 Policy Implications for the Global
South}\label{policy-implications-for-the-global-south}

These findings carry direct implications for policymakers in the Global
South. First, they underscore that sovereign compute is not about
competing at the frontier, but about securing the capacity to train
\emph{usable} models that are strategically sufficient for national and
regional needs (Sevilla et al., 2022; Sastry et al., 2024). H100-based
configurations, at 8--14 million USD, fall comfortably within a
realistic affordability envelope, while A100 deployments, at 19--32
million USD, remain viable but impose higher fiscal pressure.

Second, the relevant fiscal comparison is not with multi-billion dollar
infrastructure projects, but with digital inclusion programs. Mexico's
\emph{CFE Telecomunicaciones e Internet para Todos}, for example,
received yearly budgets of over 12,000 million pesos ($\approx$ 700 million USD)
in recent federal allocations (Comisión Federal de Electricidad, 2024).
Against this backdrop, allocating tens of millions for sovereign AI
capacity appears not only affordable but strategically prudent.

Finally, the results suggest that compute policy in the Global South
should integrate three components: (1) fiscal planning that prioritizes
efficient accelerators; (2) infrastructure investments to prepare
national grids for future scaling (SENER, 2023; Empresa de Pesquisa
Energética {[}EPE{]}, 2023); and (3) governance mechanisms to ensure
that sovereign compute is treated as a public good rather than a purely
commercial asset. Together, these elements can position emerging
economies to develop AI capabilities that are usable, transparent, and
aligned with institutional needs, without incurring the unsustainable
costs of chasing frontier AI.

\section{5. Conclusion}\label{conclusion}

This study has shown that the sovereign training of \emph{usable but
non-frontier} language models is technically and fiscally feasible for
middle-income countries. Using the case of Brazil and Mexico, we modeled
four infrastructure configurations combining H100 and A100 accelerators
across 90- and 150-day training windows. The results demonstrate that
hardware efficiency is the decisive variable: H100-based clusters fall
within 8--14 million USD and remain compatible with medium-voltage
electrical infrastructure, while A100 deployments, at 19--32 million
USD, are less fiscally viable due to their higher energy and capital
demands.

These findings highlight a central distinction for compute governance in
the Global South. Sovereign actors do not need to pursue frontier AI
systems---whose costs run into the hundreds of millions---but can
instead develop strategically sufficient models that are usable,
auditable, and aligned with national priorities. In this sense, training
time emerges as a policy lever: extending schedules allows countries to
adapt to hardware constraints without requiring immediate access to the
most advanced accelerators.

Looking ahead, the challenge will be to integrate sovereign compute into
broader strategies for digital infrastructure and energy planning.
Fiscal support for efficient accelerators, targeted investments in
medium-voltage grid capacity, and governance mechanisms that treat
compute as a public good can ensure that usable AI models are developed
in ways that serve institutional and societal needs. By focusing on
context-sensitive feasibility rather than frontier competitiveness,
countries like Brazil and Mexico can chart realistic pathways toward
technological sovereignty in the age of AI.

\section*{REFERENCES}

\begin{itemize}
\item
  ANEEL -- Agência Nacional de Energia Elétrica. (2024). \emph{Tarifas
  de Energia Elétrica -- Dados e Publicações}.
    \url{https://www.gov.br/aneel/pt-br/assuntos/tarifas}
\item
  Artificial Analysis. (2025). \emph{Model Benchmark Explorer}.
  \url{https://artificialanalysis.ai}
\item
  BIS- U.S. Bureau of Industry and Security. (2025). \emph{Brazil
  Country Commercial Guide.
  https://www.trade.gov/country-commercial-guides/brazil-import-tariffs}
\item
  CFE-Comisión Federal de Electricidad. (2024). \emph{Informe de la
  gestión gubernamental 2019--2024: CFE Telecomunicaciones e Internet
  para Todos}. CFE.
  \url{https://www.cfe.gob.mx/nuestraempresa/Informe\%20de\%20gestion\%201824/EPS\%20Telecomuniaciones\%20e\%20Internet\%20para\%20Todos.pdf}
\item
  CFE- Comisión \ul{Federal de Electricidad. (2025)} \emph{Tarifa GDMTH
  (Media Tensión Horaria)}
  \url{https://app.cfe.mx/Aplicaciones/CCFE/Tarifas/TarifasCRENegocio/Tarifas/GranDemandaMTH.aspx}
\item
  Department for Science, Innovation and Technology (DSIT). (2024, Oct).
  \emph{Frontier AI: Capabilities and risks -- Discussion paper}. UK
  Government.
  \url{https://www.gov.uk/government/publications/frontier-ai-capabilities-and-risks-discussion-paper}
\item
  \emph{Empresa de Pesquisa Energética. (2023). Plano decenal de expansão de energia 2034 (PDE 2034). Ministério de Minas e Energia.}
  \url{https://www.epe.gov.br/sites-pt/publicacoes-dados-abertos/publicacoes/PublicacoesArquivos/publicacao-804/topico-758/PDE2034_Aprovado.pdf}
\item
  Epoch AI. (2024). \emph{Tracking Large-Scale AI Models}.
  \url{https://epoch.ai/blog/tracking-large-scale-ai-models}
\item
  Gobierno de México. (2023, 15 de agosto). \emph{Decreto por el que se
  modifica la Tarifa de la Ley de los Impuestos Generales de Importación
  y de Exportación}. Diario Oficial de la Federación.
  \url{https://dof.gob.mx/nota_detalle_popup.php?codigo=5724207}
\item
  International Energy Agency (IEA). (2025). \emph{Energy and AI: How
  much power will AI use?}
  \url{https://www.iea.org/reports/energy-and-ai/energy-supply-for-ai}
\item
  NVIDIA A100 TENSOR CORE GPU:
  \url{https://www.nvidia.com/content/dam/en-zz/Solutions/Data-Center/a100/pdf/nvidia-a100-datasheet.pdf}
\item
  NVIDIA H100 TENSOR CORE GPU:
  \url{https://www.megware.com/fileadmin/user_upload/LandingPage\%20NVIDIA/nvidia-h100-datasheet.pdf}
\item
  ONS -- Operador Nacional do Sistema Elétrico (Brasil). (2009).
  \emph{Procedimentos de Rede -- Submódulo 3.6 (Estudos Elétricos para
  Acesso)}.
  \url{https://www.ons.org.br/\%2FProcedimentosDeRede\%2FM\%C3\%B3dulo\%203\%2FSubm\%C3\%B3dulo\%203.6\%2FSubm\%C3\%B3dulo\%203.6.pdf}
\item
  Patterson, D., Gonzalez, J., Le, Q. V., Liang, C., Munguía, L. M.,
  \emph{et al.} (2022). The carbon footprint of machine learning
  training will plateau, then shrink. \emph{IEEE Computer, 55}(12),
  18--28.
  \url{https://par.nsf.gov/servlets/purl/10399992}
\item
  Sastry, G., Heim, L., Belfield, H., Anderljung, M., Brundage, M.,
  Hazell, J., O'Keefe, C., Hadfield, G., Ngo, R., Pilz, K., Gor, G.,
  Bluemke, E., Shoker, S., Egan, J., Trager, R., Avin, S., Weller, A.,
  Bengio, Y., \& Coyle, D. (2024). Computing power and the governance of
  artificial intelligence. \url{https://arxiv.org/html/2402.08797v1\#S4}
\item
  SENER -- Secretaría de Energía (México). (2023). \emph{PRODESEN
  2023--2037 (Programa de Desarrollo del Sistema Eléctrico Nacional)}.
  \url{https://base.energia.gob.mx/PRODESEN2023/Capitulo1.pdf}
\item
  Sevilla, J., Heim, L., Ho, A., Besiroglu, T., Hobbhahn, M., \&
  Villalobos, P. (2022). \emph{Compute trends across three eras of
  machine learning} (arXiv:2202.05924).
  \url{https://doi.org/10.48550/arXiv.2202.05924}
\end{itemize}

\section{Annex 1.}\label{annex-1.}

\href{https://docs.google.com/spreadsheets/d/1NJ3s2nmV7fsEpphn5VulpQpYa5vAjBuO/edit?usp=sharing&ouid=106941722368252898715&rtpof=true&sd=true}{\ul{Annex 1.xlsx}}

\end{document}